\pdfoutput=1

\documentclass[11pt]{article}

\usepackage[]{ACL2023}

\usepackage{times}
\usepackage{latexsym}
\usepackage[utf8]{inputenc} 
\usepackage[T1]{fontenc}    
\usepackage{hyperref}       
\usepackage{url}            
\usepackage{booktabs}       
\usepackage{amsfonts}       
\usepackage{nicefrac}       
\usepackage{microtype}      
\usepackage[ruled]{algorithm2e}
\usepackage{threeparttable}

\usepackage{graphicx}
\usepackage{subfigure}
\usepackage{booktabs} 
\usepackage{caption}
\usepackage{framed}
\usepackage{amssymb}
\usepackage{mathrsfs}
\usepackage{mathtools}
\usepackage{array}
\usepackage{amsthm}
\usepackage{verbatim} 
\usepackage{enumerate}
\usepackage{bbm}
\usepackage{commath}
\usepackage{wrapfig}
\usepackage{amsbsy}
\usepackage{float}
\usepackage{amsmath}
\usepackage{tabularx}
\usepackage{listings}
\usepackage{xcolor}
\usepackage{colortbl}
\usepackage[shortlabels]{enumitem}
\usepackage{tcolorbox}
\usepackage{multirow}

\usepackage{algorithm}

\usepackage{algorithmic}
\usepackage[T1]{fontenc}

\usepackage[utf8]{inputenc}

\usepackage{microtype}

\usepackage{inconsolata}

\usepackage{booktabs} 
\usepackage{multirow}
\usepackage{array} 

\usepackage{amsmath}

\usepackage{graphicx} 

\newcommand{\model}{\textsc{ExpRAG}\xspace}
\newcommand{\bench}{\textsc{DischargeQA}\xspace}

\newcommand{\vpara}[1]{\vspace{0.07in}\noindent\textbf{#1 }}

\newcommand{\hide}[1]

\title{Experience Retrieval-Augmentation with Electronic Health Records Enables Accurate Discharge QA}

\author{
Justice Ou$^{1}$ \thanks{Correspondence to \texttt{zo6@illinois.edu}} \thanks{The two first authors made equal contributions.}, 
Tinglin Huang$^{2}$ \footnotemark[2],   
Yilun Zhao$^{2}$ ,   
Ziyang Yu$^{3}$,  
Peiqing Lu, 
Rex Ying$^{2}$\\
  University of Illinois Urbana-Champaign$^{1}$,   
  Yale University$^{2}$ , 
  University of Waterloo$^{3}$  \\\ 
}

\begin{document}
\maketitle
\begin{abstract}
\

To improve the reliability of Large Language Models (LLMs) in clinical applications, retrieval-augmented generation (RAG) is extensively applied to provide factual medical knowledge.
Beyond general medical knowledge, clinical case-based knowledge is also critical for effective medical reasoning, as it provides context grounded in real-world patient experiences.
Motivated by this, we propose Experience Retrieval-Augmentation~(\model) framework based on Electronic Health Record~(EHR), aiming to offer the relevant context from other patients' discharge reports.
\model performs retrieval through a coarse-to-fine process: it first applies an EHR-based report ranker to efficiently identify similar patients as experience, and then utilizes a context retriever to extract task-relevant content for enhanced medical reasoning.
To evaluate RAG systems on EHR data including \model and medical agents, we introduce \bench, a clinical QA dataset with 1,280 discharge-related questions across diagnosis, medication, and instruction tasks. 
Each problem is generated using historical EHR data to ensure realistic and challenging scenarios.
Experimental results demonstrate that \model consistently outperforms traditional text-based rankers, achieving an average relative improvement of 5.2\%, highlighting the importance of case-based knowledge for medical reasoning.

\end{abstract}

\section{Introduction}

\begin{figure}[t]
    \centering
    \includegraphics[width=0.5\textwidth,  clip]{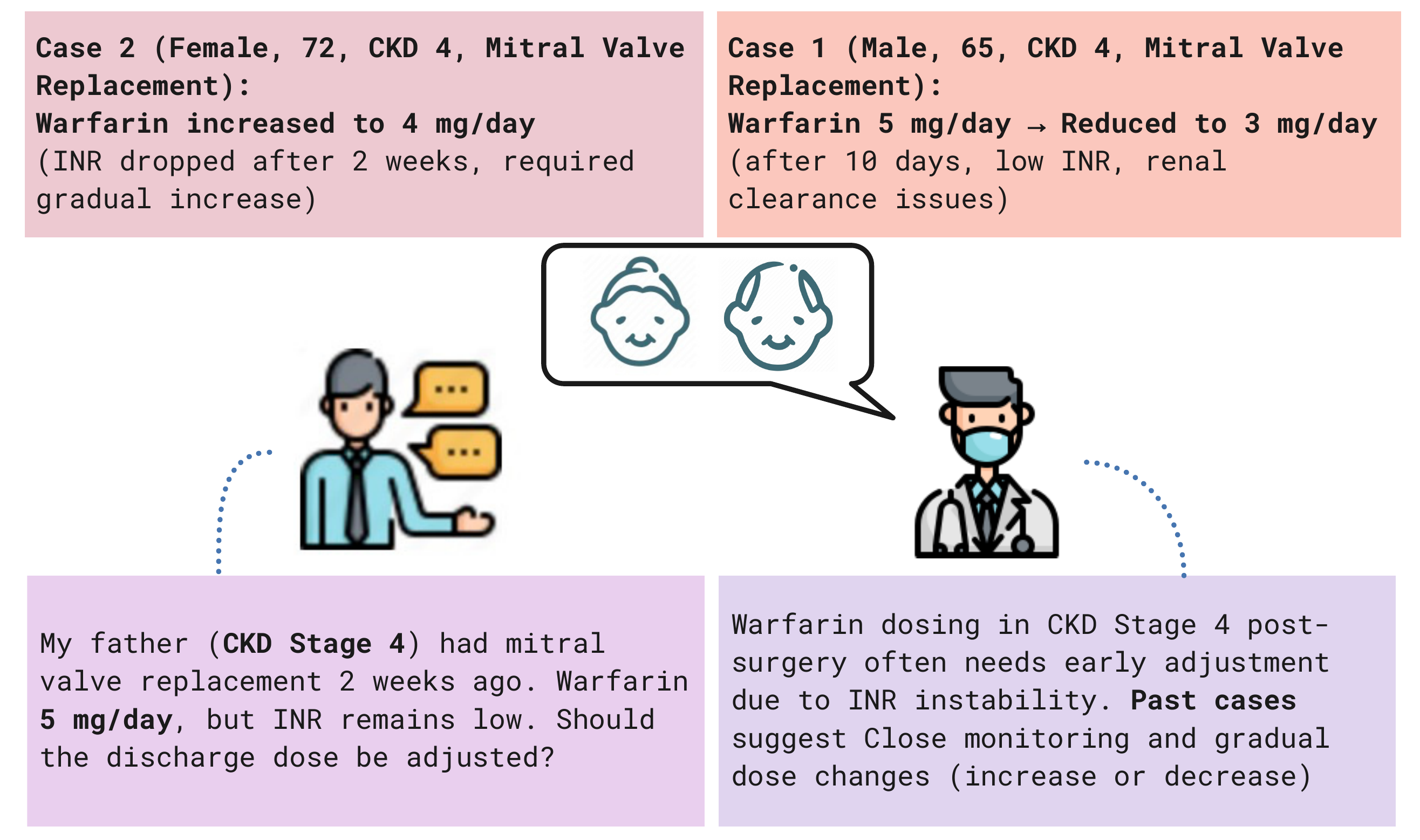}
    \caption{An illustrative example of utilizing experience from relevant clinical cases to support medical decision: adjusting a patient's warfarin dosage based on the specific clinical context rather than relying on a generic standard dose}
    \label{past_cases}
\end{figure}

Benefiting from pretraining on large-scale corpora, Large Language Models (LLMs) are capable of performing complex reasoning and have shown great promise in medical applications~\cite{zheng2024large,liu2024survey}.
One important application is inferring clinical conditions, including diagnosis and medication, which can be formulated as a question-answering~(QA) task~\cite{singhal2025toward,chen2023meditron70b,huang2024heart}.
However, LLM agents often suffer from hallucinations and a lack of domain-specific knowledge, which limits their reliability in real-world medical applications.

To address this, prior studies have resorted to retrieving factual knowledge from open-ended databases to provide context, such as the description of drugs from Wikipedia~\cite{xiong-etal-2024-benchmarking,yang2025retrieval}.
Such external knowledge enables LLMs to access general medical facts, thereby improving response accuracy.
However, introducing such general facts cannot effectively help LLMs solve real clinical cases, which often involve coexisting clinical conditions.
For example, as shown in Figure~\ref{past_cases}, adjusting a patient's warfarin dosage requires reasoning based on the specific clinical context, whereas conventional retrieval can only provide the standard dosage for warfarin, which is irrelevant in this case.

In light of this, we argue that, in addition to general factual concepts, clinical case-based knowledge is also crucial for effective medical reasoning.
The intuition is that an experienced clinician often relies on past cases with similar conditions to guide diagnosis, treatment decisions, and discharge planning.
To this end, we propose \textbf{Experience Retrieval-Augmentation~(\model)} framework, leveraging a large-scale EHR database MIMIC-IV~\cite{johnson2023mimic} as its knowledge basis.
Specifically, \model breaks down the retrieval process into two steps: (1) report ranking applies an EHR-based similarity measurement to identify patients with similar medical conditions, and (2) experience retrieval extracts problem-relevant content from these patients' discharge reports, which serves as the case-based contextual knowledge for LLMs.
The introduced EHR modality enables large-scale clinical experience retrieval, grounding the model's reasoning in real-world clinical practices.

To evaluate capability of \model and other RAG methods/agents in medical reasoning, we introduce \bench, a clinical dataset including 1,280 QA pairs dedicated to discharge-related problems.
The dataset primarily includes three types of problems:simulating the discharge process of  final diagnosis, medication prescription and post-discharge instructions. 
For each problem, we follow the data structure of the discharge report and select the content preceding the question as the problem background to avoid label leakage.
Additionally, we index the option candidates using EHR to generate contextually relevant options, ensuring a non-trivial and clinically meaningful challenge for the model.

We evaluate the performance of five different LLMs and compare \model with the text-based report ranker using \bench. 
\footnote{Uploading to \url{https://physionet.org/} 
 , please check updates on Github: \url{https://github.com/jou2024/EXPRAG}}

The results demonstrate the effectiveness of using EHR to retrieve relevant clinical experience, as it consistently improves the performance of LLM backbones and outperforms the text-based ranker with an average relative improvement of 5.2\%.
Our main contributions are summarized as follows:
\begin{itemize}[leftmargin=*]
\item We propose \model, an EHR-based experience retrieval-augmentation framework, shedding light on the potential of leveraging past clinical cases to enhance LLM performance in medical reasoning tasks.
\item We introduce \bench, a medical QA dataset for discharge-related questions, designed to evaluate LLMs' ability to simulate the clinical decision-making process during patient discharge with a more challenging setup.
\item Our results demonstrate the advantage of \model over the text-based ranker, highlighting the effectiveness of EHR in providing clinically meaningful context.
\end{itemize}

\begin{figure*}[t]
    \centering
    \includegraphics[width=1\textwidth,  clip]{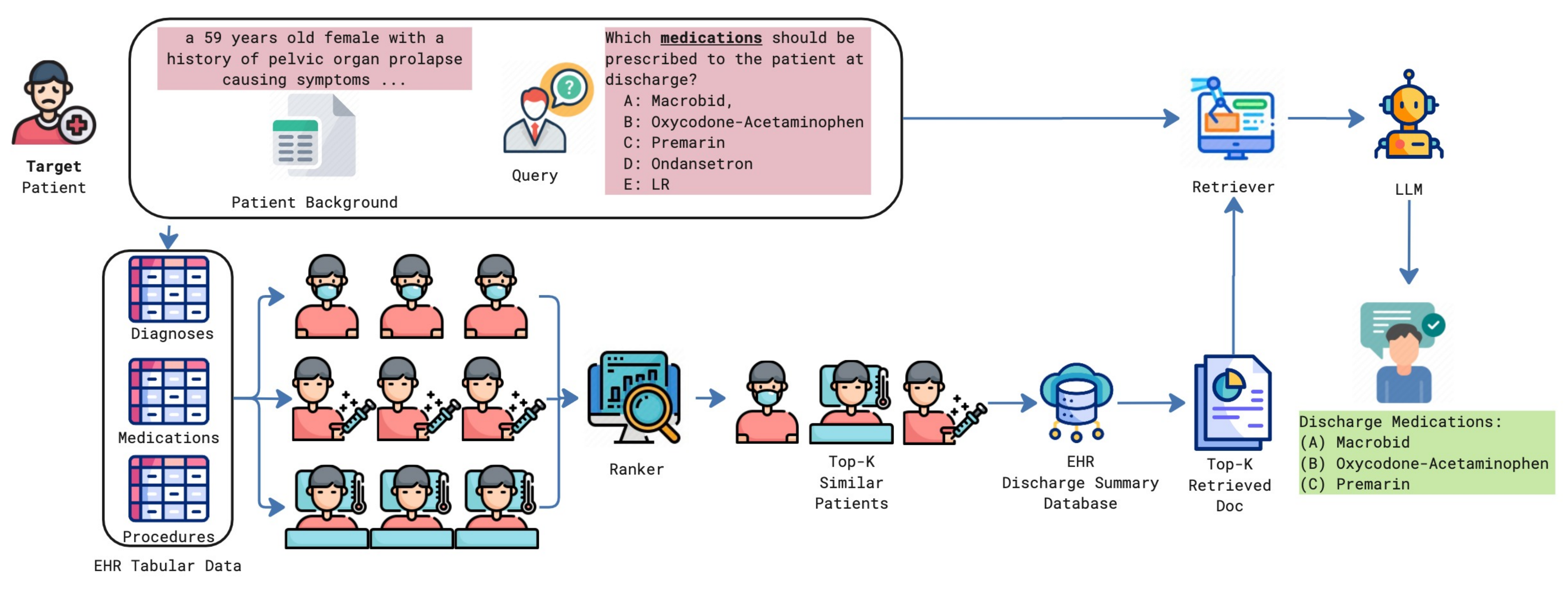}
    \caption{The overview of \model: Given a medical query and the patient's background, \model first indexes similar patients based on diagnosis, medication, and procedure similarity from the EHR.
    A text retriever is then applied to the discharge reports of the top-ranked similar patients to extract clinically relevant content, which is subsequently fed into the LLM to generate the answer.
    }
    \label{based-ranker}\label{fig:workflow}
\end{figure*}

\section{Related Work}

\vpara{Retrieval-Augmented Generation (RAG).} 
RAG has become a key paradigm for overcoming the static knowledge limitations of LLMs by retrieving external information~\cite{gu2018search, petroni2019language}. 
Traditional RAG frameworks typically use dense retrieval methods to augment generative tasks~\cite{DBLP:conf/naacl/DevlinCLT19, DBLP:conf/iclr/XiongXLTLBAO21}. While effective in general QA, these methods often lack domain specificity, which is critical in healthcare~\cite{lu-etal-2024-clinicalrag}.
Recent advancements like ClinicalRAG~\cite{lu-etal-2024-clinicalrag} and MIRAGE~\cite{xiong-etal-2024-benchmarking} address this by integrating structured EHR data and clinical notes for diagnosis and treatment planning. However, existing benchmarks primarily focus on isolated information retrieval~\cite{johnson2023mimicivnote, kweon2024ehrnoteqa}, overlooking the complexities of reasoning over patient histories and similar cases. To bridge this gap, our work extends RAG by combining structured EHR data with discharge summaries, enabling experience-driven reasoning for more realistic and reliable medical QA.

\vpara{Medical QA Benchmark.}
EHRSQL~\citep{lee2022ehrsql} and DrugEHRQA~\citep{bardhan-etal-2022-drugehrqa} target structured data queries, with the former addressing SQL-based operations and the latter focusing on drug-related questions.
EHRNoteQA~\citep{kweon2024ehrnoteqa} and RadQA~\citep{soni-etal-2022-radqa} leverage clinician-verified QA pairs from discharge summaries and radiology reports, while MedQA~\citep{jin2021disease}, MedMCQA~\citep{pal2022medmcqa}, and PubMedQA~\citep{jin2019pubmedqa} evaluate LLMs with questions from medical exams or PubMed articles.
Discharge-summary-focused datasets like emrQA~\citep{pampari2018emrqa} and CliniQG4QA~\citep{yue2021cliniqg4qa} use discharge notes for QA tasks, and specialized datasets like RxWhyQA~\citep{fan2019annotating} and drug-reasoning QA~\citep{moon2023extractive} focus on specific question types like medication reasoning.
Unlike these benchmarks, \bench introduces an evaluation framework centered around the discharge process, simulating the clinical workflow from diagnosis inference to medication prescription and discharge instruction generation.
Additionally, we leverage EHR to generate non-trivial, contextually relevant candidate options, providing a more challenging and realistic setup.

\section{\model Framework}
\model provides a comprehensive framework for retrieving relevant knowledge from the cohort, as shown in Figure~\ref{fig:workflow}.
In this section, we first formulate the problem that \model aims to tackle and then elaborate the two-step retrieval framework.

\subsection{Task Formulation}
A cohort contains a set of discharge report $\mathcal{D}=\{D_i\}_N$ where $D_i=\{d_j\}_M$ denotes the $i$-th report and $d_j$ is $j$-th paragraph in $D_i$.
Each report is a medical document that offers an overview of a patient's hospitalization. 
The goal of \model is to extract relevant content from $\mathcal{D}$ that helps LLM to effectively answer a given medical query $q$ related to a specific patient $p$:
\begin{align}
   d_{*}=f_{\model}(p,q,\mathcal{D})
\end{align}
The queries studied in this work focus on providing professional medical guidance for patients, including diagnosis, medication, and discharge instructions, thereby simulating realistic and practical clinical scenarios, as discussed in Section~\ref{sec:dataset}.

Different from the conventional RAG focusing on extracting factual concepts from open-ended databases, \model aims to utilize contextually-relevant clinical practice, inspired by how doctors collect and apply experience from past clinical cases.
These two approaches rely on different reasoning procedures and knowledge sources, making them complementary to each other.

\subsection{Coarse-to-Fine Retrieval Framework.}
To retrieve information from the cohort, one naive solution is to concatenate all the reports into one document and apply a text retriever to extract relevant content, similar to the conventional RAG pipeline. 
However, a standard EHR cohort typically contains millions of hospital visits, making it impractical to exhaustively search over all reports. 


To efficiently perform experience retrieval, \model applies a two-step framework which conduct the retrieval from a coarse to fine level:

\vpara{Report Ranking.}
Before addressing a specific medical query, an intuitive assumption is that only patients with similar clinical histories, e.g., similar diseases or medications, can potentially provide meaningful guidance. 
In light of this, \model first employs a report ranker to efficiently discard unrelated cases and narrow down the candidate pool using the patient information:
\begin{align}
   \mathcal{D}'=f_{\text{Ranker}}(p,\mathcal{D})
\end{align}
where $\mathcal{D}'=\{D_i\}_{N'\ll N}$ is a small subset of the selected discharge summaries.
The ranker module will be scalable and enable effective utilization of patient context.
Specifically, we introduce EHR as a knowledge base to facilitate the patient-level similarity measurement, as presented in Section~\ref{sec:ehr_ranker}.


\vpara{Experience Retrieval.} 
Based on the selected candidate pool, a sophisticated text retriever is capable of providing more accurate and dedicated clinical experience searching:
\begin{align}
   d_{*}=f_{\text{Retriever}}(q,\mathcal{D}')
\end{align}
Built on top of clinically relevant reports identified by the dedicated ranking approach, the retriever focuses on extracting content related to the medical query. 
We here apply existing text retrievers, such as auto-merging or BM25, during this phase.

\subsection{EHR-Based Report Ranker $f_{\text{Ranker}}$}\label{sec:ehr_ranker}
Electronic Health Record~(EHR), as a structured data organization,
typically consists of multiple tabular data, each recording specific medical information about patients.
In this study, we focus on measuring the similarity between patients using the following three medical entities:
\begin{itemize}[leftmargin=*]
\item Diagnosis: Identified disease assigned to a patient, represented by ICD-10 code.
\item Medication: Prescribed drugs administered to a patient, recorded using NDC code.
\item Procedure: Medical intervention, operation, or clinical process performed on a patient, represented by ICD-10 code.
\end{itemize}
Quantify the similarity between patients based on these three dimensions offers a comprehensive criterion for identifying clinically relevant reports.

For a patient $p$, these three medical entities are represented as sets $E^{\text{Diag}}_p$, $E^{\text{Med}}_p$, and $E^{\text{Proc}}_p$, respectively.
Given two patients $p$ and $p'$, we first compute the set similarity between them using each medical information:
\begin{align}\label{equ:balancing}
   \tau_{\text{Diag}}&=f_\text{similarity}(E^{\text{Diag}}_p,E^{\text{Diag}}_{p'}),\\
   \tau_{\text{Med}}&=f_\text{similarity}(E^{\text{Med}}_p,E^{\text{Med}}_{p'}),\\
   \tau_{\text{Proc}}&=f_\text{similarity}(E^{\text{Proc}}_p,E^{\text{Proc}}_{p'})
\end{align}
where $f_\text{similarity}(\cdot,\cdot)$ is a set similarity metric, with the Jaccard Index applied in this study.
Finally, these similarity metrics are aggregated using a weighted sum:
\begin{align}
\tau=\lambda_1\tau_{\text{Diag}}+\lambda_2\tau_{\text{Med}}+\lambda_3\tau_{\text{Proc}}
\end{align}
where $\lambda_{1/2/3}$ is the hyperparameter balancing the importance of each metric.
We perform pairwise similarity comparisons between the query patient and other patients within the EHR, returning the discharge summaries of the top-$k$ most similar patients as results.

\vpara{Efficiency Analysis.}
The overall computation is practically efficient since the computation of Jaccard Index can be significantly accelerated with some libraries, such as Faiss~\cite{douze2024faiss} and NumPy~\cite{harris2020array}.
Besides, indexing medical entities from tabular data enables fast lookups, further reducing computational overhead.

\begin{table*}[ht!]
\centering
\scriptsize
\setlength{\tabcolsep}{4pt}
{\fontsize{6pt}{7pt}\selectfont
\renewcommand{\arraystretch}{1.2}
\begin{tabular}{@{} m{1.3cm}m{0.9cm}m{1.3cm}m{0.75cm}m{2.5cm}m{3cm}m{1.75cm}m{1.5cm} @{}}
\toprule
      & \textbf{Task} & \textbf{Response Type} & \#\textbf{Query}  &  \textbf{Example Question}      &  \textbf{Practical Significance}        &  \textbf{Background Source}    &  \textbf{Option Source}\\ \midrule
     & Diagnosis    Inference    & Multi-select      & 436       & "Which diagnoses should be documented in the patient's discharge summary?"    & Reflects a doctor’s process of identifying all relevant diagnoses based on clinical profile.         & Clinical profile  & Discharge Report    \& EHR \\ \cline{2-8} 
\multirow{3}{*}{\textbf{\bench}}      & Medications      Inference    & Multi-select      & 444       & "Which medications should be prescribed to the patient at discharge?"         & Simulates the doctor's task of ensuring correct medications are prescribed based on hospital treatment. & Clinical profile \& In-hospital progress & Discharge Report    \& EHR \\ \cline{2-8} 
      & Instructions      Inference   & Single-select     & 400       & "What is the best instruction for this patient?"       & Mimics the final step of doctors' advising patients with appropriate post-discharge care instructions.  & Clinical profile \& In-hospital progress & Discharge    Report \& AI \\ \midrule
EHRNoteQA & Clinical Inference & Single-select \& Open-Ended  & 962       & "What was the treatment provided for the patient’s left breast cellulitis?" & Extract and answer based on content from full discharge notes        & Full clinical notes    & Discharge    Report \& AI     \\ \midrule
CliniQG4QA         & Retrieval & Text Span         & 1,287      & "Why has the patient been prescribed hctz?"  & Retrieve the related content as answer from report & Full clinical notes & /\ \\ \bottomrule
\end{tabular}

\caption{\label{tab:qa_benchmarks} \centering
Comparison of \bench and the previous EHR-related QA benchmarks.
}
}
\vspace{5mm}
\end{table*}


\section{\bench Dataset}\label{sec:dataset}
To evaluate LLMs' capability in utilizing the retrieved experience, we construct a medical question-answering dataset specifically designed for discharge-related queries based on MIMIC-IV~\cite{johnson2023mimic}.
Each question in the dataset pertains to critical discharge information, including the patient's diagnosis, prescribed medications, and post-discharge instructions.

\subsection{Dataset Introduction}

\vpara{Overview.}
As shown in Table~\ref{tab:qa_benchmarks}, \bench consists of a total of 1,280 QA pairs, each associated with a patient ID and corresponding clinical background. 
The questions in \bench can be categorized into three main types:
\begin{itemize}[leftmargin=*]
\item Diagnosis Inference: Questions related to identifying the patient’s medical diagnosis.
\item Medication Inference: Questions regarding the medications prescribed, including dosage, frequency, and purpose.
\item Instruction Inference: Questions focused on discharge instructions, such as follow-up care, activity restrictions, and self-care guidelines.
\end{itemize}
These three categories collectively cover the key aspects of discharge-related patient care, requiring LLMs to perform non-trivial reasoning based on the given clinical background.  
Moreover, all these problems can potentially benefit from the retrieved experience, mimicking the way clinicians apply past clinical knowledge to make medical decisions during discharge.


\begin{figure}[t]
    \centering
    \includegraphics[width=0.5\textwidth,  clip]{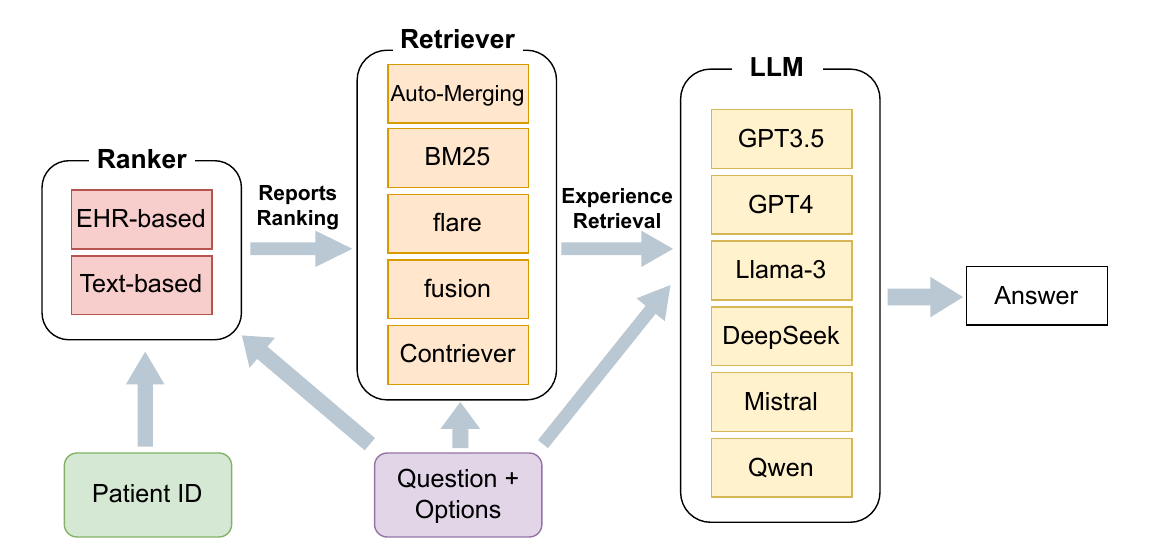}
    \caption{Inference pipeline of \bench.}
    \label{fig:Inference_pipeline}
\end{figure}

\vpara{Evaluation Settings.}
Each problem in \bench includes a problem description, the patient's clinical background, and multiple options for the LLM to choose from.
While instruction inference uses a single-select setup, diagnosis and medication inference adopt a multi-select setup, requiring the model to select multiple options to answer the questions. Each option corresponds to a specific diagnosis or treatment.
This multi-select format presents a more challenging and realistic setting for LLMs, as clinicians often need to identify and address multiple coexisting conditions.

The overall inference pipeline is presented in Figure~\ref{fig:Inference_pipeline}, where we implement several components, including Ranker, Retrieval, and LLM agent, to support discharge-related QA using \model.

\vpara{Comparison.}

Compared with existing QA benchmarks that focus primarily on general clinical QA or information retrieval, \bench centers on the discharge procedure, simulating a doctor's medical reasoning process: inferring the diagnosis from the clinical profile, prescribing appropriate treatments, and summarizing the condition—offering a more realistic scenario.
Additionally, we utilize EHR to generate contextually relevant options, requiring LLMs to perform non-trivial reasoning to solve the tasks.
More details are provided in Table~\ref{tab:qa_benchmarks}.

\begin{table*}[!t]
\centering
\small
\begin{tabular}{@{}llccccc@{}}
\toprule
\multirow{2}{*}{\textbf{Model}} & \multirow{2}{*} {\textbf{Context}} & \multicolumn{1}{l}{\textbf{Instruction}} & \multicolumn{2}{c}{\textbf{Diagnosis}} & \multicolumn{2}{c}{\textbf{Medication}} \\ \cmidrule(l){3-7} 
                 &                  &\textbf{Acc}(\%)      &\textbf{Acc}(\%) & \textbf{F1}    &\textbf{Acc}(\%)        & \textbf{F1}      \\ \midrule
\multirow{3}{*}{Mistral-7b}       
& Direct-Ask                  & 67.0               & 16.03  & 0.488                  &  1.13   & 0.362               \\
& bge-small-en                & \textbf{70.0}      & 14.67  & 0.490                  & 0.90  & 0.356               \\
& \model$_{\text{EHR}}$       & 69.0               & 13.79 & \textbf{0.505}          & 1.13  & \textbf{0.371}             \\ 
\midrule
\multirow{3}{*}{Deepseek-R1-8B}      
& Direct-Ask                   & 70.0              & 10.78 & 0.363              & 0.69 & 0.217               \\
& bge-small-en                 & 72.0              & 12.84  & \textbf{0.381}    & 1.58 & 0.230               \\
& \model$_{\text{EHR}}$        & \textbf{75.3}     & 11.01 & 0.379              & 0.9 & \textbf{0.241}              \\ 
\midrule
\multirow{3}{*}{Qwen3-30B-A3B}       
& Direct-Ask                   & 90.8             & 15.60  & 0.415                      & 1.13 & 0.280               \\
& bge-small-en                 & 93.8             & 18.12 & 0.502                       & 2.25  & \textbf{0.366}       \\
& \model$_{\text{EHR}}$        & \underline{\textbf{95.3}} & 17.43   & \textbf{0.528}   & 1.35  & 0.355             \\ 
\midrule
\multirow{3}{*}{GPT-3.5}          
& Direct-Ask                  & 73.0             & 18.50 &  0.498             & 1.15 & 0.234             \\
& bge-small-en                & 78.8             & 15.60 &  0.405             & 0.68 & 0.317              \\
& \model$_{\text{EHR}}$       & \textbf{79.5}    & 18.81 &  \textbf{0.504}    & 1.80 & \textbf{0.371}              \\
\midrule
\multirow{3}{*}{GPT-4o}           
& Direct-Ask                  & 90.0             & 9.86 & 0.510               & 3.65 & 0.486             \\
& bge-small-en                & 90.3             & 8.26 & 0.493               & 4.95 & 0.601             \\
& \model$_{\text{EHR}}$       & \textbf{91.3} & \underline{\textbf{21.33}} & \underline{\textbf{0.530}}  & \underline{\textbf{9.68}} & \underline{\textbf{0.638}} \\
\bottomrule
\end{tabular}

\caption{\label{tab:method_results} Performance Comparison Across Multiple LLMs using \bench}

\end{table*}

\subsection{Dataset Construction}

\vpara{Patients Filtering.}
\label{patients_filtering}
We first filter out low-quality patient records in MIMIC-IV for various reasons.
Starting with all 430,000 patients, we remove encounters without available discharge summaries, leaving 320,000 patients.
Next, we filter out patients with fewer than 3 or more than 40 entries in any of the diagnosis, medication, or procedure records, resulting in a final dataset of 28,000 patients for generating QA pairs. 
For instruction inference, we further exclude patients with excessively short discharge summaries using GPT-4o, as explained in \ref{app:dischargeQA_construction}.

\vpara{Background Generation.}
To enable LLMs to make realistic medical decision, it is necessary to offer a clinical background of the patient along with the question.
To \textbf{avoid label leakage} during the context generation, we propose leveraging the structured format of the discharge summary, which consists of the following components:
\begin{itemize}[leftmargin=*]
\item Clinical profile: Essential patient demography, the presenting condition, and initial clinical assessments.
\item In-hospital progress: The interventions, therapies, and the patient's clinical progress during hospitalization.
\item Discharge plan summary: The details of discharge diagnosis and medication, and instructions for post-hospital care.
\end{itemize}
An illustrative example can be found in Appendix~\ref{app:dataset}.
These three components represent the core of discharge decision-making procedure, aligning with the three main problem types in \bench.
Accordingly, we present the summarized sections preceding the questions as context. For example, the clinical profile serves as the background for diagnosis-related questions, while the in-hospital progress is additionally included for medication-related questions.
Notably, basic patient demographic information is always included as part of the contextual background.

\vpara{Option Generation.}
For diagnosis and medication inference, we utilize EHR to generate non-trivial candidate options by extracting all associated diagnoses and medications of a patient and feeding them into GPT-4o to select contextually relevant candidates, ensuring a challenging selection process. 
For instruction inference, GPT-4o first summarizes the key points of the ground-truth answers and then applies permutations to generate plausible yet incorrect candidate options. Notably, EHR tabular data usually has \textbf{very limited overlap} with our task for discharge, such as in-hospital medications, due to professionalism and use cases, have much limitation than discharge prescriptions.  For all tasks, the correct options are directly extracted from the discharge reports.

\section{Experiments}

In this section, we evaluate \model on \bench with five LLMs, comparing it with the text-based report rankers.
We also analyze the effects of balancing coefficients, the number of similar patients, and retriever choices, followed by case studies on the retrieved experience.

\subsection{Comparison of LLMs}
We evaluated the performance of 4 state-of-the-art LLMs of varying scales, ranging from close-source to 8 billion parameters open-source model, on three clinical tasks: discharge instructions, diagnosis, and medications. These models included GPT-3.5~\cite{openai2022chatgpt}, GPT-4o~\cite{openai2024gpt4o}),  Mistral-7B~\cite{Jiang2023Mistral7}, and two thinking models—Deepseek-R1-8B~\cite{guo2025deepseek} and Qwen3-30B-A3B~\cite{qwen3}.

\vpara{Hyperparameters.}
We also present the results of LLMs with \model. 
The default balancing coefficients $\lambda_{1/2/3}$ are set to 1/3 each and auto-merging~\cite{LlamaIndex} is used as the retriever.
The number of similar patients is set as 15.

\vpara{Metrics.} We report accuracy across all the tasks, calculated as the percentage of correctly answered questions.
Note that for multi-select problems, a question is considered answered correctly only when all correct options are selected.
We additionally report F1 scores for the multi-select problems to provide a more comprehensive analysis of LLM performance on these challenging tasks.

\vpara{Results.}
Table~\ref{tab:method_results} summarizes the performance of all LLMs with \model on the three clinical tasks. GPT-4o consistently achieves the best performance across tasks, with an absolute improvement of 13\% over GPT-3.5 on instruction inference problems.
Additionally, the multi-select tasks (diagnosis and medication) prove significantly more challenging, as most LLMs achieve accuracy below 20\%, indicating the limited medical reasoning capabilities of current large language models, the value of challenge of \bench.


\subsection{Comparison of Report Ranker}

To verify the effectiveness of \model and the utilization of EHR, we implement baselines that perform report ranking solely based on text, i.e., a text-based ranker, as a key part in other traditional RAG methods.
Using embedding model \texttt{bge-small-en-v1.5} \cite{bge_embedding}, a popular model due to its light weight and high correlation scores in Table~\ref{tab:retrieval}, we embed queries (question, options, background) to be computed similarity with each discharge report embedding.
The top-$k$ similar reports are then retrieved, followed by a text retriever to extract the relevant information.

The results are presented in Table~\ref{tab:method_results}.
We observe that \model outperforms the text-based ranker in most cases, achieving an average relative improvement of 5.2\%.
Notably, the EHR-based ranker leverages structured EHR data for ranking, eliminating the need for an embedding process and thereby enabling a more efficient pipeline.

%
%
%

As shown in Table~\ref{tab:reranker_compare}, we compare our EHR‐based ranker with two additional sentence-embedding models: \texttt{all-MiniLM-L6-v2} and \texttt{paraphrase-MiniLM-L3-v2} \cite{reimers-2019-sentence-bert}, using GPT3.5 as backbone LLM.  
As a result, \model$_{\text{EHR}}$ outperforms all text-based variants in terms of all the metrics. 
\begin{table}[!t]
\centering
\small

\renewcommand{\arraystretch}{1.2} 
\setlength{\tabcolsep}{4pt}      

\begin{tabular}{@{}l ccccc @{}}
\toprule
\multirow{2}{*}{\textbf{Rankers}} 
& \multicolumn{1}{c}{\textbf{Instruction}} 
& \multicolumn{2}{c}{\textbf{Diagnosis}} 
& \multicolumn{2}{c}{\textbf{Medication}} \\ 
\cmidrule(l){2-6} 
& \textbf{Acc} 
& \textbf{Acc} 
& \textbf{F1} 
& \textbf{Acc} 
& \textbf{F1} \\ 
\midrule
bge-small-en          & 78.8  & 15.60  & 0.405 & 0.68  & 0.317  \\
all-MiniLM-L6           & 79.3  & 17.43  & 0.476 & 1.13  & 0.316  \\
paraphrase-L3    & 77.5  & 18.81  & 0.492 & 1.58  & 0.330  \\ 
\model$_{\text{EHR}}$                 & \textbf{79.5}  & \textbf{18.81}  & \textbf{0.504} & \textbf{1.80}  & \textbf{0.371}  \\ 
\bottomrule
\end{tabular}

\caption{\label{tab:reranker_compare} Performance Comparison with Embedding Models as Reranker vs EHR-based Reranker.}

\end{table}

\begin{table}[!t]
\centering
\small

\renewcommand{\arraystretch}{1.2} 
\setlength{\tabcolsep}{4pt}      

\begin{tabular}{@{} l c c @{}}
\toprule
\textbf{Rankers} & \textbf{Pearson} & \textbf{Spearman} \\
\midrule
bge-small-en-v1.5           & 0.639  & 0.623 \\
all-MiniLM-L6            & 0.640  & 0.618 \\
paraphrase-MiniLM-L3     & 0.478  & 0.481 \\
\model$_{\text{EHR}}$           & \textbf{0.669}  & \textbf{0.648} \\
\bottomrule
\end{tabular}

\caption{\label{tab:retrieval} Retrieval Performance Comparison of Rerankers using Pearson and Spearman Correlation.}

\end{table}
\vpara{Retrieval Correlation Comparison.}
We conducted an experiment comparing patients similarity scores generated by the EHR-based method with those reliably annotated by LLMs. A higher correlation between the two indicates stronger retrieval performance by the EHR-based approach.
Specifically, we randomly sample 100 target patients from \bench, as detailed in Appendix~\ref{app:reranker-details}. Each target–candidate pair is scored by GPT4o-mini on the three modalities (diagnoses, procedures, prescriptions) with a single “overall” similarity. For each ranker (i.e., three text-embedding models and our EHR-based \model), we compute Pearson and Spearman correlation between its scores and annotations, and average over all 100 targets (Table~\ref{tab:retrieval}).  Over the strongest text-embedding baselines,the results by \model confirm that explicit EHR similarity provides more faithful retrieval signals than generic sentence embeddings.


\begin{figure}[!htb]
    \centering
    \includegraphics[width=0.5\textwidth,  clip]{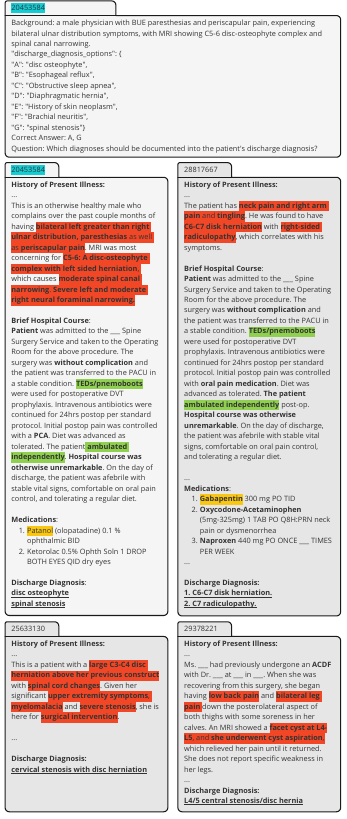}
    \caption{Comparison of Similar Patients.}
    \label{fig:case_study}
\end{figure}
\subsection{Further Analysis}

We conducted additional studies to investigate the impact of key components on the performance of \model, including the number of similar patients $k$, and the balancing coefficients $\lambda_{1/2/3}$.
GPT-3.5 is applied as the backbone by default.

\begin{table}[!t]
\centering
\small
\renewcommand{\arraystretch}{1.2} 
\setlength{\tabcolsep}{3pt}       

\begin{tabular}{@{}p{1.9cm} ccccc @{}}
\toprule
\multirow{2}{*}{\textbf{Model}} 
& \multicolumn{1}{c}{\textbf{Instruction}} 
& \multicolumn{2}{c}{\textbf{Diagnosis}} 
& \multicolumn{2}{c}{\textbf{Medication}} \\ 
\cmidrule(l){2-6} 
& \textbf{Acc} 
& \textbf{Acc} 
& \textbf{F1} 
& \textbf{Acc} 
& \textbf{F1} \\ 
\midrule
Uniform                 & 79.5  & 18.81 & 0.504 & 1.8  & 0.371  \\
Task-focused         & 77.5  & 10.91  & 0.377 & 0.91  & 0.322  \\
Complementary         & 76.8  & 18.18  & 0.446 & 2.73  & 0.305  \\ 
\bottomrule
\end{tabular}

\caption{\label{tab:balance} Performance Comparison for Coefficients Distributions \bench}
\end{table}

\vpara{Balancing Coefficients.}
As shown in Equ.~\ref{equ:balancing}, we introduce three coefficients to balance the similarity computed based on diagnosis, medication, and procedures. 
We apply an equal distribution by default and explore the effect of using different weighting strategies here:
\begin{itemize}[leftmargin=*]
\item Task-focused weighting: Assign a weight of 1 to the task-relevant similarity measure and 0 to the others. For example, $\lambda_1=1,\lambda_2=0,\lambda_3=0$ for diagnosis inference.
\item Complementary weighting: Assign a weight of 1 to the two less relevant similarity measures while setting the task-relevant measure to 0. For example, $\lambda_1=0,\lambda_2=1,\lambda_3=1$ for diagnosis inference.
\end{itemize}
The results are shown in Table~\ref{tab:balance}. We can find that complementary weighting can achieve the best performance in most cases, demonstrating that information from multiple clinical dimensions can provide a more comprehensive context.

\vpara{Top-$k$ Patients.}
We vary the number of retrieved similar patients $k$ on QA performance. As shown in Table~\ref{tab:topk_all}, different tasks have varies trends. For example, on Instruction task, accuracy starts to increases with larger number.  
While for Diagnosis and Medication tasks, beyond $k=20$, performance fluctuates, suggesting that while more retrieved candidates provide useful context, excessive retrieval may introduce irrelevant or conflicting information, leading to slight declines in accuracy.

\begin{table}[!t]
\centering
\small

\renewcommand{\arraystretch}{1.1} 
\setlength{\tabcolsep}{4pt}      

\begin{tabular}{@{}p{2.0cm} ccccc @{}}
\toprule
\multirow{2}{*}{\footnotesize \# Similar Patients} 
& \multicolumn{1}{c}{\textbf{Instruction}} 
& \multicolumn{2}{c}{\textbf{Diagnosis}} 
& \multicolumn{2}{c}{\textbf{Medication}} \\ 
\cmidrule(l){2-6} 
& \textbf{Acc} 
& \textbf{Acc} 
& \textbf{F1} 
& \textbf{Acc} 
& \textbf{F1} \\ 
\midrule
$k=5$         & 80.00  & 19.04  & 0.511 & 1.80  & 0.366  \\
$k=10$        & 78.75  & 18.58  & 0.511 & 1.35  & 0.371  \\
$k=15$        & 79.50  & 18.81  & 0.504 & 1.80  & 0.371  \\ 
$k=20$        & 80.75  & 19.50  & 0.524 & 1.80  & 0.377  \\ 
$k=25$        & 82.25  & 19.27  & 0.515 & 1.35  & 0.352  \\ 
\bottomrule
\end{tabular}

\caption{\label{tab:topk_all} GPT-3.5 performance with different $k$.}

\end{table}

\subsection{Case Study}
We perform a focused case study on a Discharge Diagnosis task from \bench. Specifically, we analyze a patient (ID: 20453584) presenting with bilateral ulnar paresthesias and neck pain. Similar patients are identified by matching ICD/NDC codes from structured EHR data (Figure~\ref{fig:case_study}). Reviewing discharge summaries of these similar patients revealed shared key diagnostic features—including cervical disc herniation, spinal stenosis, and upper extremity neurological symptoms—which substantially clarified the target patient’s clinical picture. For instance, similar patients exhibiting C6-C7 disc herniations and spinal stenosis provided critical evidence, improving the interpretation of the target patient's symptoms and supporting a more accurate final diagnosis. We elaborate the explanation in  Appendix~\ref{app:case_study_details}.

\section{Conclusion}  
Inspired by the importance of experience in clinical decision-making, we propose a novel coarse-to-fine retrieval framework, \model, to utilize knowledge from similar patient records.
Specifically, we introduce EHR as a knowledge basis and employ a reliable similarity measurement algorithm to narrow down the candidate pool to have relevant and useful content. 
Evaluated on our curated \bench, \model consistently improves the performance of various LLMs, highlighting the potential of leveraging past experience to enhance model performance on medical QA.

\vpara{Limitations}
While EHR provides abundant medical information, such as lab test results, our proposed \model currently utilizes only diagnosis, medication, and procedures as an initial exploration, which provides valuable insights and promising directions for future research.
Additionally, \bench currently consists solely of multi-option questions, which can be enhanced to be open-ended to comprehensively evaluate the generative capabilities of LLMs, When more economical and accurate LLMs are developed.


\bibliography{anthology,custom,llm, rag_related}

\appendix
\clearpage
\section{Appendix}

\begin{figure*}[!htbp]
    \centering
    \includegraphics[width=1\textwidth,  clip]{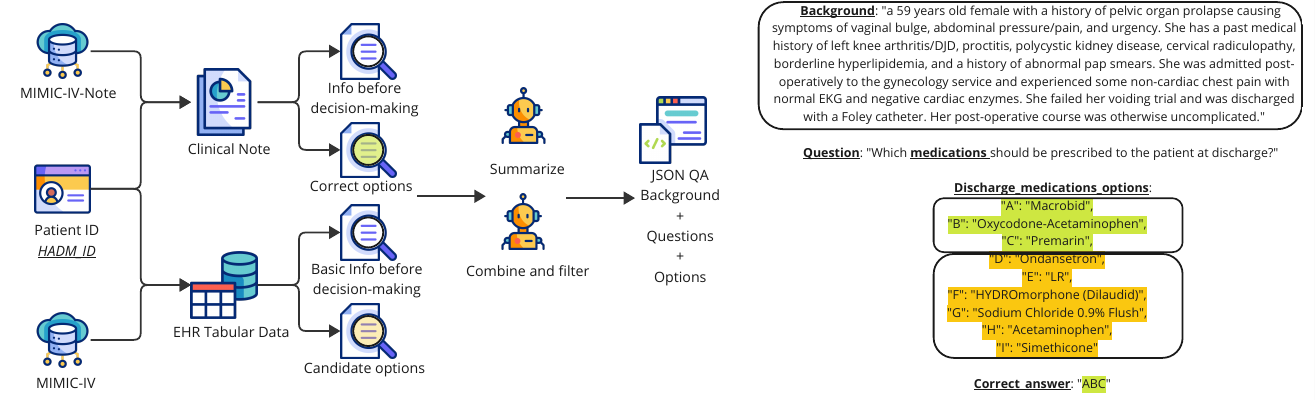}
    \caption{\label{fig:qa_construct} DischargeQA generation workflow.}
\end{figure*}

\subsection{\bench}
\label{app:dischargeQA_construction}
To evaluate LLMs in real‐world clinical scenarios, we constructed a dataset combining structured tables from MIMIC-IV and unstructured discharge notes from MIMIC-IV-note, totaling over 140 000 patient records. 
The whole process is shown in \ref{fig:qa_construct}

\paragraph{Patient selection.}
As in ~\ref{patients_filtering}, a valid pool of candidates for target or similar patients is the key point for retrieving information from similar patients.

Starting from the 430 k patients in MIMIC-IV, we retain
(i) encounters with a discharge summary,
(ii) 3–40 rows in each of \texttt{diagnoses}, \texttt{prescriptions}, and \texttt{procedures},
and (iii) discharge notes that contain \(\ge 4\) instruction bullet points identified by GPT-4o.
This yields 28 000 admissions, each can give enough structured items and narrative content to form challenging QA pairs, so that there are enough information to compare to other patients, and to have candidate options as wrong answer if there is no overlap in discharge reports.

\paragraph{Note segmentation and exposure.}
Each discharge summary is heuristically split into seven sections
\emph{(Patient Demography, Presenting Condition, Clinical Assessment, Treatment Plan, In-Hospital Progress, Discharge Summary, Post-Discharge Instructions)}
and mapped to three temporal phases: \textbf{pre-diagnosis}, \textbf{in-hospital}, and \textbf{post-discharge}, matching the 3 sections \textbf{Clinical profile}, \textbf{In-hospital}, and \textbf{Discharge Plan} as Figure ~\ref{fig:clinical_notes}. 
For every task we reveal only the phases that would have been available to the clinician at decision time, preventing label leakage.

\paragraph{Problem Design with EHR}
\begin{itemize}
  \item \textbf{Golden answers} are clinician‐authored discharge–note items.
  \item \textbf{Distractors} are drawn from the same patient’s structured tables (drugs for medication, ICD-coded diagnoses for diagnosis).
  \item Overlaps between structured candidates and golden answers are merged via GPT-4o.
  \item Unlike single-choice formats, both diagnosis and medication tasks use \emph{multi-select}, requiring selection of \emph{all} correct items, thus raising task difficulty.  
\end{itemize}

\paragraph{Construction Prompt Design}
Prompts to generate options are carefully crafted to combine EHR tabular data and golden answer from discharge note. As an example, the prompt as Figure \ref{fig:prompt-diag} defines the role of the model as a clinician and provides three key components: the list of discharge diagnoses, a database of historical diagnoses, and summarized background information extracted from the discharge summary. The task requires the model to identify which diagnoses should be included in the discharge summary by reasoning through the given data. Correct options are derived from the discharge diagnosis list, while incorrect but plausible options are generated from the diagnoses database. To ensure realism, GPT-4o is used to handle overlap, summarize long diagnoses, and align outputs with clinical expectations, providing a rigorous framework for evaluation.

\subsection{Compare Two Rankers using similar patients}\label{app:ranker_compare}
\begin{figure*}[!htbp]
    \centering
    \includegraphics[width=1\textwidth,  clip]{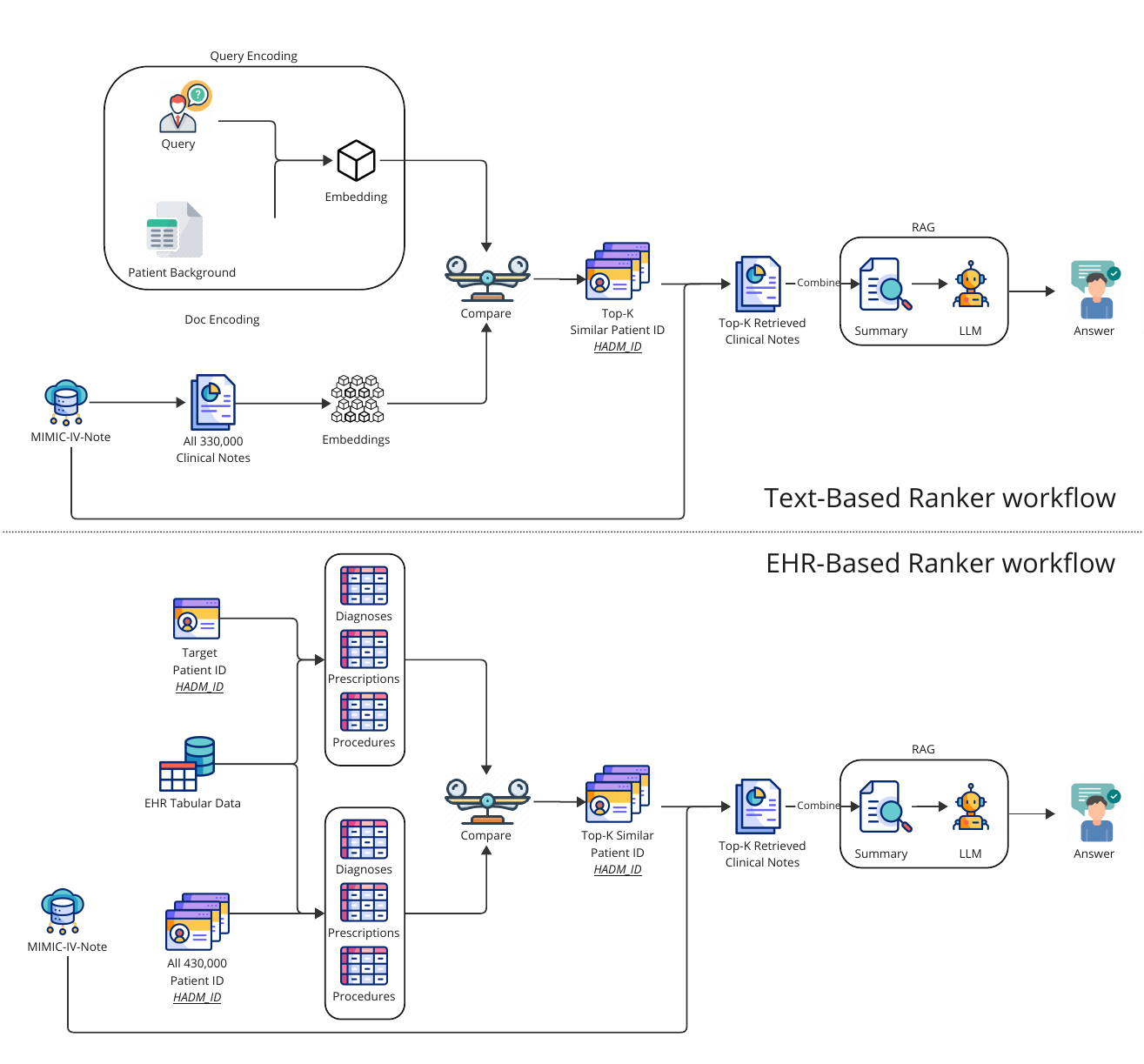}
    \caption{\label{fig:ranker_compare} Text-Based Ranker vs EHR-Based Ranker workflow.}
\end{figure*}

Shown as Figure ~\ref{fig:ranker_compare}. 
\subsubsection{Conventional RAG}
A traditional RAG pipeline for clinical question answering consists of:
\begin{enumerate}
\item Encoding a query (e.g., a discharge planning question) into an embedding vector.
\item Chunking a large corpus into smaller text passages and embedding each chunk.
\item Finding top-k relevant text chunks by comparing similarity scores between query embeddings and chunk embeddings.
\item Retrieving the most relevant text passages.
\item Feeding the retrieved text to the LLM for answer generation.
\end{enumerate}
However, applying this approach to EHR discharge summaries presents challenges:
\textbf{Inefficiency}: The MIMIC-IV discharge summary corpus is over 4GB in size, making fine-grained chunk retrieval computationally expensive.
\textbf{Loss of Context}: Traditional chunking disrupts the continuity of patient history, making it difficult for LLMs to infer longitudinal medical decisions.
\textbf{Scattered Information:} Important information may be spread across multiple notes, making individual paragraph retrieval suboptimal.

\subsubsection{Text-Based Ranker}
Instead of retrieving individual paragraphs or sentences, we treat each patient's discharge summary as a retrievable document and adapt the RAG pipeline accordingly:

\vpara{Query Encoding} 
Convert the question, options and target patient's background into an embedding vector.

\vpara{Patient-Level Indexing} 
Store each patient’s discharge summary as a separate retrievable document and compute its embedding.

\vpara{Find Similar Patients Using Embeddings} 
Rank top-k similar patients from all patients based on similarity between the query embedding and patient embeddings.

\vpara{Retrieve Top-K Patient Summaries} Select the top-k most similar patient discharge summaries as a document source, like "experience". 

\vpara{Run RAG on Retrieved Summaries} Use the retrieved summaries as references for LLM-based answer generation. Traditional RAG is running by embedding the query and the document source, retrieving chunks as reference, and answering query to generate the correct answer.

By preserving full discharge summaries, Text-based retrieval maintains patient-level coherence while improving retrieval efficiency.

\subsubsection{EHR-Based Ranker}
To further improve retrieval relevance, we introduce an EHR-based Ranker that utilizes structured patient data for retrieving better discharge summaries as the "experience" document source:

\vpara{Identify the Target Patient’s Condition} Extract structured EHR tabular data (ICD codes for diagnoses, prescriptions, and procedures) from the target patient.

\vpara{Find Similar Patients Using Structured EHR Data} Compute similarity between the target patient’s structured data and all other patients based on shared ICD codes in MIMIC-IV, after scanning all recorded patient data. Jaccard index is used to calculate the similarity. 

\vpara{Retrieve Top-K Similar Patients} Select the top-k most similar patients based on their structured medical records.

\vpara{Extract Their Discharge Summaries} Retrieve the corresponding discharge summaries for the top-k similar patients.

\vpara{Run RAG on Retrieved Summaries} Feed the retrieved summaries into an LLM for answer generation, same as the last step in Text-Based Ranker.

By incorporating structured EHR data, this method ensures that retrieval is clinically relevant, going beyond semantic text similarity. This process resembles an experience-based search: identifying similar past experiences and adapting them to solve the problem at hand.

\subsection{Dataset Details}\label{app:dataset}

\vpara{MIMIC-IV\footnote{\url{https://mimic.mit.edu/docs/iv/}}} This database~\cite{johnson2016mimiciii,johnson2023mimic,johnson2023mimicivnote} contains information on over 40,000 patients admitted to the critical care units at Beth Israel Deaconess Medical Center from 2001 to 2012 and has been widely used in prior research.
The database is publicly available for research purposes, with strict de-identification protocols to protect patient privacy, making it a valuable resource for developing and evaluating machine learning models in healthcare.
The data is hierarchically organized, with each patient record comprising multiple encounters, each containing various entities such as demographics, medications, diagnoses, procedures, and lab results.
Additionally, the database includes unstructured data, such as discharge reports and X-ray images, with each admission marked by a date and timestamp.

\begin{figure*}[t]
    \centering
    \includegraphics[width=1\textwidth,  clip]{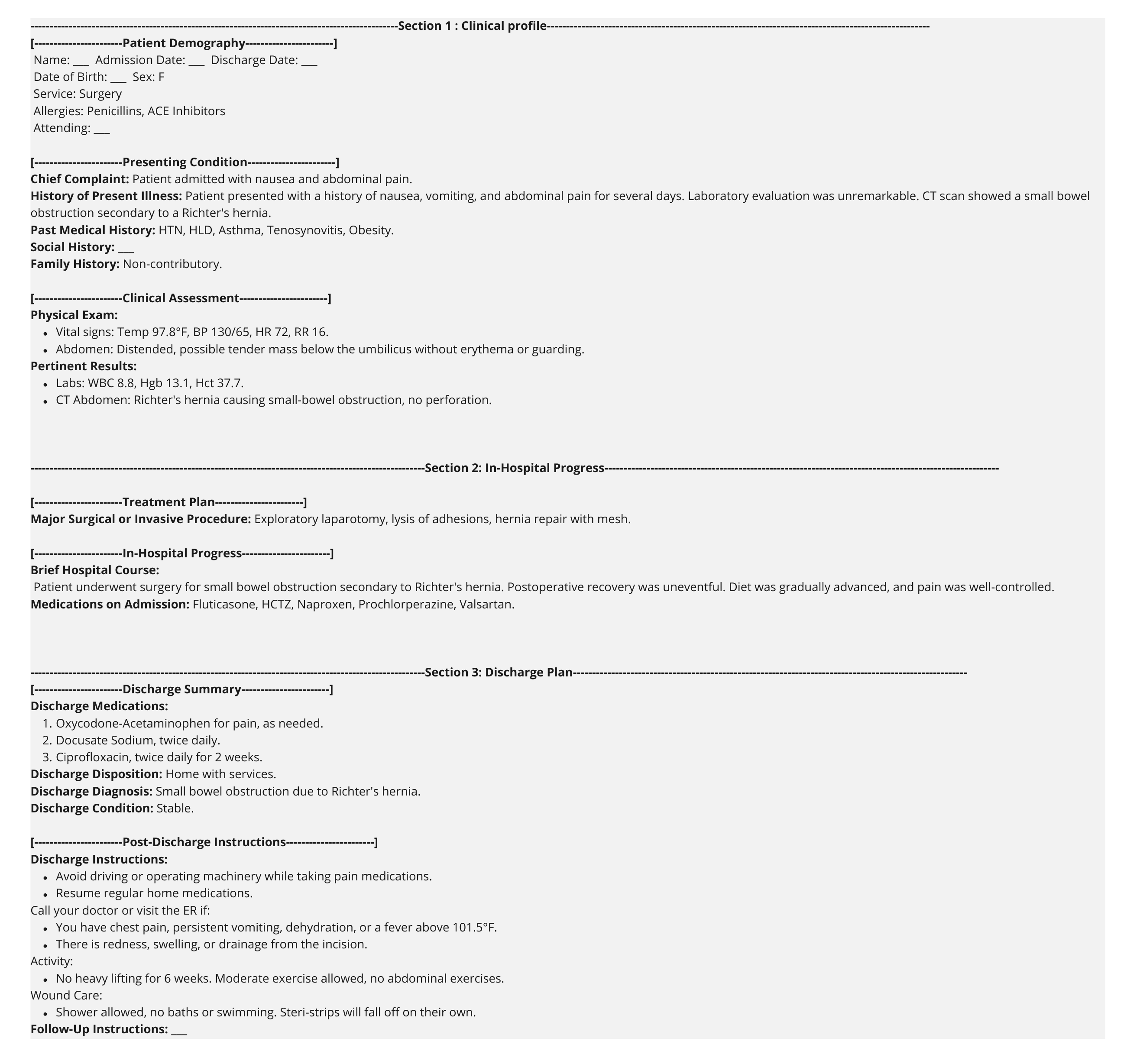}
    \caption{\label{fig:clinical_notes} An example of discharge report, which can be split into 3 sections and 7 subsections: clinical profile, in-hospital progress, and discharge plan summary. 
    Note: Some pertinent results from exams are before diagnosis, while some are after diagnosis or after procedures.}
\end{figure*}

\vpara{Structure of Discharge Report}
As shown in Figure~\ref{fig:clinical_notes}, a discharge summary follows a structured format that provides a comprehensive overview of a patient’s hospitalization and care journey. It is typically divided into three main parts: (1) The Clinical Profile, which includes essential patient information, presenting conditions, and the initial clinical assessment upon admission; (2) The In-Hospital Progress, which documents the treatment plan, administered therapies, and the patient’s progress throughout the hospital stay; and (3) The Discharge Plan Summary, which summarizes the patient's discharge status, prescribed medications, and detailed post-discharge instructions for ongoing care. 

\subsection{Retriever}\label{app:retriever}
\begin{table}[!t]
\centering
\small

\renewcommand{\arraystretch}{1.2} 
\setlength{\tabcolsep}{4pt}      

\begin{tabular}{@{}l c@{}}
\toprule
\textbf{Retriever}
& \textbf{Instruction Acc} \\ 
\midrule
Auto Merging    & 79.5 \\
Sentence Window & 74.5 \\
BM25            & 68.0 \\
BM25+          & 69.0 \\
Flare           & 74.5 \\
Contriever      & 69.0 \\
\bottomrule
\end{tabular}

\caption{\label{tab:retrievers} Performance Comparison of Retrievers.}

\end{table}
\subsubsection{Retriever Experiments}
\label{app:exp_retriever}
Our experiments include: Auto-merging~\cite{LlamaIndex}, sentence-window~\cite{LlamaIndex}, BM25~\cite{robertson2009probabilistic}, BM25+~(the combination of BM25 and word embeddings), Contriever~\cite{izacard2021unsupervised}, and flare~\cite{jiang2023active}. 
More details regarding the methods can be found in Appendix~\ref{app:retriever}.
As shown in Table~\ref{tab:retrievers}, auto-merging, sentence-window, and flare achieve the best performance, highlighting the effectiveness of context-aware retrieval.
However, Contriever, which utilizes unsupervised dense representations, also underperforms in our cases, suggesting the need for medical domain-specific fine-tuning.

\vpara{Auto-merging}
Auto-merging retrieval in RAG by LlamaIndex \cite{LlamaIndex} hierarchically structures documents into parent and child nodes, allowing for the retrieval of larger, more coherent context by merging child nodes into parent nodes when multiple related chunks are relevant to a query

\vpara{Sentence-window}
Sentence-window retrieval by LlamaIndex \cite{LlamaIndex} parses documents into individual sentences with surrounding context, enabling fine-grained retrieval while maintaining local coherence by including a window of adjacent sentences

\vpara{BM25 \& BM25+}
BM25 (Best Match 25) is a ranking function used in information retrieval that scores documents based on the frequency of query terms within them, taking into account document length and term frequency saturation.
BM25+ by LlamaIndex \cite{LlamaIndex} combines retrieval methods BM25 and vector-based retrieval, to leverage the strengths of both approaches. This hybrid technique allows for capturing both keyword relevance and semantic similarity, often using algorithms like Relative Score Fusion to re-rank and merge results from different retrievers, resulting in more accurate and comprehensive search outcomes.

\vpara{Flare}
FLARE (Forward-Looking Active REtrieval) enhances RAG by enabling the language model to anticipate future content needs, iteratively predicting upcoming sentences and retrieving relevant information when encountering low-confidence tokens, thus improving response accuracy and contextual relevance

\vpara{Contriever}
Contriever is a single-tower dense retrieval model that employs self-supervised contrastive learning to enhance document embeddings for retrieval tasks. It encodes both queries and documents using the same encoder, producing dense vector representations. The model utilizes a self-supervised contrastive learning approach with a loss function that optimizes embeddings by comparing relevant passages to negative (irrelevant) ones

\subsection{LLM Backbone}

\subsubsection{Proprietary Models} 
We selected GPT-3.5 and GPT-4o as representative models due to their extensive real-world adoption and practical relevance demonstrate the applicability of our framework in everyday clinical settings. 

\vpara{GPT3.5 \& GPT-4o}
GPT-3.5, developed by OpenAI and released in November 2022, was followed by GPT-4o, also created by OpenAI and launched on May 13, 2024, marking a significant advancement with its ability to process and generate outputs across text, audio, and image modalities in real time.

\subsubsection{Open-source Models}
Qwen3, Deepseek-R1 and Mistral were included as leading open-source models to benchmark the generalizability and robustness of our approach. Specifically,  Qwen3 and Deepseek-R1 are thinking models, which is an essential feature to tackle challenges in our tasks. 

\vpara{Deepseek-R1-8B}
DeepSeek-R1 is a powerful 671 billion parameter language model developed by DeepSeek AI, from which we use DeepSeek-R1-Distill-Llama-8B was derived as a more efficient 8 billion parameter version, distilled from the original model's knowledge to offer improved performance on reasoning tasks while maintaining computational efficiency

\vpara{Mistral}
Mistral-7B-Instruct-v0.3 is an advanced instruction-tuned language model featuring an extended vocabulary of 32,768 tokens, support for the v3 Tokenizer, and function calling capabilities, enabling more versatile and complex interactions compared to its predecessors

\vpara{Qwen3-30B-A3B}
Qwen3-30B-A3B is developed by Alibaba Cloud, released at April 2025. It is a dense-and-MoE model that has a “thinking mode” for deep reasoning, math, and coding. It outperforms earlier Qwen generations on logical reasoning, code generation, tool-using agent tasks, and human-preference benchmarks, while supporting 100 + languages for instruction following and translation. In our work we leverage its thinking mode to maximize accuracy on complex reasoning tasks. 

\subsubsection{Medical Models}
\label{medical_llms}
\vpara{Baichuan-M1}
Trained from scratch on 20 T tokens that mix high-quality clinical and general texts, Baichuan-14B-M1 is the first open-source 14 B-parameter LLM purpose-built for medicine. It incorporates specialised heads, enabling fine-grained reasoning that matches general-domain peers on standard benchmarks yet surpasses models 5× larger on medical tasks. An updated architecture with longer-context handling further improves comprehension of lengthy clinical narratives and complex patient histories.

We have experiments shown in ~\ref{medical_llms}. Baichuan-14B-M1 exhibits the same limitations we observed in other compact models. When additional context from EHR- or text-based retrieval is supplied, the model occasionally confuses the target patient with the retrieved similar cases. Under RAG settings, this confusion produces > 5 \% invalid responses—outputs that either fail our answer-parsing patterns or omit a decisive answer—leading to a noticeable overall drop in accuracy. Despite these invalid cases, Baichuan-M1 using EHR-Based \model still surpasses other approaches on the Medication task, underscoring its strength in pharmacological reasoning even when other aspects falter.

\begin{table}[!t]
\centering
\small

\renewcommand{\arraystretch}{1.2} 
\setlength{\tabcolsep}{2.8pt}      

\begin{tabular}{@{}llccc@{}}
\toprule
\multirow{2}{*}{\textbf{Model}} & \multirow{2}{*} {\textbf{Context}} & \multicolumn{1}{l}{\textbf{Instruction}} & \multicolumn{1}{c}{\textbf{Diagnosis}} & \multicolumn{1}{c}{\textbf{Medication}} \\ \cmidrule(l){3-5} 
                 &                  &\textbf{Acc}(\%)      & \textbf{F1}            & \textbf{F1}      \\ \midrule
\multirow{3}{*}{Baichuan}       
& Direct-Ask       & 89.8               & 0.381                    & 0.256               \\
& Text-based       & 87.8               & 0.377                    & 0.277               \\
& \model$_{\text{EHR}}$        & 86.3               & 0.373           & \textbf{0.285}             \\ 
\bottomrule
\end{tabular}

\caption{\label{tab:medical_llms} Performance from Biomedical Model, with 5\% invalid answer on Text-based and EHR-based}

\end{table}

\vpara{UltraMedical}
Llama-3-8B-UltraMedical is derived from Meta’s Llama-3-8B and further tuned on the 410 K-example UltraMedical dataset (synthetic + curated), an 8 B-parameter biomedical specialist. It tops MedQA, MedMCQA, PubMedQA and MMLU-Medical, handily beating larger general models such as GPT-3.5 and Meditron-70B, and rivals domain-tuned Flan-PaLM and OpenBioLM-8B. The model targets exam-style question answering, literature comprehension and clinical-knowledge retrieval, making advanced medical NLP accessible at modest compute cost.

UltraMedical’s 70 B variant has a 8 k-token context window. When we integrated the model into our retrieval-augmented generation pipeline, more than 50 \% of the assembled prompts—target patient record + similar-patient excerpts + task instruction—were longer than 8 k tokens. These over-length inputs had to be truncated or rejected, which systematically stripped clinical details from half of our queries and degraded answer quality. Because this hard context limit blocked reliable end-to-end evaluation, we discontinued UltraMedical-70B for our study despite its otherwise strong domain results.

\subsection{Prompt Templates}
\begin{figure*}[t]
    \centering
    \includegraphics[width=1\textwidth,  clip]{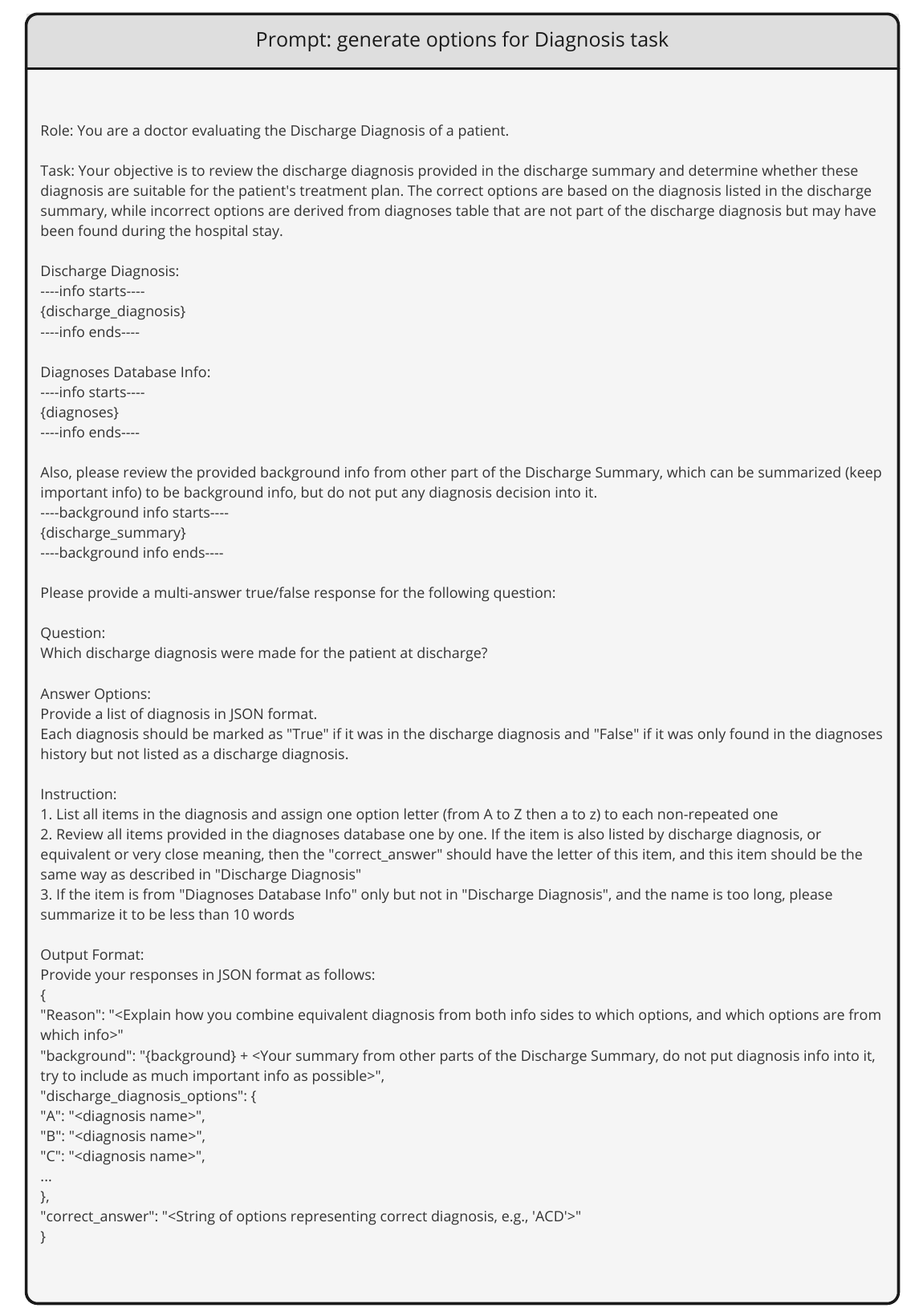}
    \caption{\label{fig:prompt-diag} Prompt design example for Diagnosis tasks}
\end{figure*}

Figure~\ref{fig:prompt-diag} presents the prompt used to generate Diagnosis task options, which is  the same way as generating options for Medication task. 

\subsection{Detailed Setup for Ranker Evaluation}
\label{app:reranker-details}

\paragraph{Patient and Candidate Sampling.}
We first randomly select 100 target patients from the MIMIC-IV cohort. For each target, we construct 100 candidate patients by:
\begin{itemize}
  \item Sampling 20 uniformly at random from the filtered step described in \ref{patients_filtering} from full dataset (to include unrelated cases).
  \item Sampling 80 from a restricted pool formed by taking the patients (per target) with non-zero EHR similarity in either of three modalities (diagnoses, procedures, prescriptions).
\end{itemize}

\paragraph{Annotation with GPT4o-mini.}
Each target–candidate pair is judged by GPT4o-mini along the three modalities. We average these three scores to obtain a single “ground truth” similarity for correlation.

\paragraph{Ranking Methods.}
\begin{enumerate}
  \item \textbf{Text embeddings:} Cosine similarity over discharge summaries using three pretrained models (\texttt{bge-small-en-v1.5}, \texttt{all-MiniLM-L6-v2}, \texttt{paraphrase-MiniLM-L3-v2}).
  \item \textbf{EHR-based EXPRAG:} Jaccard similarity on code sets (\texttt{ICD/NDC}) for each modality, aggregated with equal weights.
\end{enumerate}

\paragraph{Correlation Metrics.}
For each ranker and target patient, we produce a ranked list of 100 candidates. We then compute Pearson’s $\rho$ and Spearman’s $\tau$ between the ranker’s scores and the GPT4o-mini annotations, and average each metric across all 100 targets.

\paragraph{Results.}
Table~\ref{tab:retrieval} (main text) reports the final average correlations, demonstrating that EHR-based EXPRAG best aligns with human-like judgments.

\subsection{Details in Case Studies}
\label{app:case_study_details}
We conduct a case study focusing on Discharge Diagnosis in \bench. We examine the discharge diagnosis of a target patient (ID: 20453584) who presented with bilateral ulnar paresthesias and neck pain. We compare this patient with similar patients, who are selected by comparing ICD/NDC codes from EHR tabular data.  Fig\ref{fig:case_study} shows one example question in \bench and the similarities in the discharge reports between the target patient and the similar patients with IDs 25633130, 29378221 and 28817667. 
Upon reviewing the discharge summaries of the similar patients, it became clear that several key diagnostic features were shared with patient 20453584:
\begin{itemize}[leftmargin=*]
    \item \textbf{Disc Herniation}: Both the target patient and similar patients had disc herniations, with the target patient experiencing a C5-6 disc-osteophyte complex and the similar patients exhibiting C3-C4 and C6-C7 herniations.
    \item \textbf{Spinal Stenosis}: Many of the similar patients displayed \textbf{spinal stenosis}, which was consistent with the target patient's symptoms of narrowing of the spinal canal and foraminal narrowing.
    \item \textbf{Upper Extremity Symptoms}: The target patient reported \textbf{bilateral ulnar paresthesias}, which mirrored the bilateral symptoms observed in several similar patients, such as neck pain radiating to the arms and tingling in the extremities.
\end{itemize}

\textbf{Results:}
By comparing the discharge summaries, key features from similar patients that influenced the diagnosis of the target patient:
\begin{itemize}
    \item Similar patients with \textbf{C6-C7 disc herniation} and \textbf{radiculopathy} helped to refine the target patient’s diagnosis, suggesting that similar nerve root involvement could explain the upper extremity symptoms.
    \item The presence of \textbf{spinal stenosis} and \textbf{neural foraminal narrowing} in several patients guided the understanding of the target patient's potential nerve compression, which contributed to the diagnosis of \textbf{spinal stenosis}.
\end{itemize}

The comparison to similar patients led to a more precise discharge diagnosis for the target patient, which included a C5-6 disc-osteophyte complex with associated \textbf{spinal canal narrowing} and \textbf{neural foraminal narrowing}. These insights allowed the LLMs to confirm the target patient's diagnosis, which aligned with options \textbf{A} and \textbf{G} — \textbf{disc osteophyte} and \textbf{spinal stenosis}.

\end{document}